\documentclass{article}

\usepackage{microtype}
\usepackage{graphicx}
\usepackage{subfigure}
\usepackage{booktabs}
\usepackage{hyperref}

\usepackage{amsmath}
\usepackage{amssymb}
\usepackage{xspace}
\usepackage{xcolor}
\usepackage{multirow}
\usepackage{makecell}
\usepackage{enumitem}
\usepackage[normalem]{ulem}
\usepackage[accepted]{mlsys2025}
\usepackage{cleveref}
\usepackage{xurl}

\newif\ifdraft

\drafttrue %

\renewcommand{\emph}[1]{\textit{#1}}
\newcommand{\MyPara}[1]{\noindent{\textbf{#1}}~}

\newcommand{\sys}{SparseSpec\xspace}
\newcommand{\sspec}{sparse self-speculative decoding}

\newcommand{\alg}{PillarAttn}

\newcommand{\kvc}{KV-Cache}

\newcommand{\lm}{Llama}

\newcommand{\sota}{state-of-the-art}

\newcommand{\fig}[1]{Figure ~\ref{#1}}
\newcommand{\refsec}[1]{\S~\ref{#1}}
\newcommand{\reftable}[1]{Table ~\ref{#1}}

\newcommand{\dl}{drafting length}

\newcommand{\ar}{acceptance rate}

\newcommand{\ubs}{unified and resource-aware batch scheduler}
\newcommand{\UBS}{Unified and Resource-aware Batch Scheduler}

\newcommand{\dkvm}{dynamic \kvc{} manager}

\newcommand{\uptocomparedtovllm}{2.13$\times$}

\newcommand{\uptocomparedtospec}{1.36$\times$}

\mlsystitlerunning{Accelerating Large-Scale Reasoning Model Inference with Sparse Self-Speculative Decoding}

\begin{document}

\twocolumn[
\mlsystitle{Accelerating Large-Scale Reasoning Model Inference: Self-Speculative Decoding with Sparse Attention}

\mlsyssetsymbol{equal}{*}

\begin{mlsysauthorlist}
\mlsysauthor{Yilong Zhao}{ucb,equal}
\mlsysauthor{Jiaming Tang}{mit,equal}
\mlsysauthor{Kan Zhu}{uw}
\mlsysauthor{Zihao Ye}{nv}
\mlsysauthor{Chi-Chih Chang}{cornell}
\mlsysauthor{Chaofan Lin}{thu}
\mlsysauthor{Jongseok Park}{ucb}
\mlsysauthor{Guangxuan Xiao}{mit}
\mlsysauthor{Mohamed S. Abdelfattah}{cornell}
\mlsysauthor{Mingyu Gao}{thu}
\mlsysauthor{Baris Kasikci}{uw}
\mlsysauthor{Song Han}{mit,nv}
\mlsysauthor{Ion Stoica}{ucb}
\end{mlsysauthorlist}

\mlsysaffiliation{ucb}{University of California, Berkeley}
\mlsysaffiliation{mit}{Massachusetts Institute of Technology}
\mlsysaffiliation{uw}{University of Washington}
\mlsysaffiliation{nv}{NVIDIA}
\mlsysaffiliation{cornell}{Cornell University}
\mlsysaffiliation{thu}{Tsinghua University}

\mlsyscorrespondingauthor{Yilong Zhao}{yilongzhao@berkeley.edu}

\vskip 0.3in
\begin{abstract}
Reasoning language models have demonstrated remarkable capabilities on challenging tasks by generating elaborate chain-of-thought solutions. 
However, such lengthy generation shifts the inference bottleneck from compute-bound to memory-bound. 
To generate each token, the model applies full attention to all previously generated tokens, requiring memory access to an increasingly large \kvc{}.
Consequently, longer generations demand more memory access for every step, leading to substantial pressure on memory bandwidth.

To address this, we introduce \sys{}, a speculative decoding framework that reuses the \textit{same model} as both the draft and target models (i.e., self-speculation). 
\sys{} features a novel sparse attention mechanism, \alg{}, as the draft model, which accurately selects critical tokens via elegantly reusing information from the verification stage. 
Furthermore, \sys{} co-designs self-speculation with three system optimizations: 
(1) a \textit{unified scheduler} to batch both draft and verification phases to maximize parallelism, 
(2) \textit{delayed verification} for CPU/GPU overlap, and 
(3) \textit{dynamic \kvc{} management} to enable host memory offload to maximize GPU memory utilization.
Across various models and datasets, \sys{} outperforms \sota{} solutions, with an up to \uptocomparedtovllm{} throughput gain. 
Code is open-sourced at \href{https://github.com/sspec-project/SparseSpec/tree/main}{github.com/sspec-project/SparseSpec}.

\end{abstract}
]

\printAffiliationsAndNotice{\mlsysEqualContribution}

\section{Introduction}
\label{sec:intro}

Recent advances in reasoning language models (RLMs), such as OpenAI-o1~\cite{openai2024openaio1card}, have demonstrated remarkable capabilities in solving complex reasoning tasks.
These models typically generate \textit{tens of thousands of tokens} from problems described in only hundreds of tokens through extensive chain-of-thought (CoT)~\cite{snell2024scalingllmtesttimecompute,deepseekai2025deepseekr1incentivizingreasoningcapability}. 
This lengthy, deliberate reasoning paradigm shifts the performance bottleneck of inference from compute-bound to \textit{memory-bound}~\cite{zhao2024blendserveoptimizingofflineinference}.
Due to the auto-regressive nature of RLMs, generating each token requires loading all previously generated key-value vectors (\kvc{}), making long-output tasks memory-bound. 
In fact, the total amount of \kvc{} that needs to be loaded increases \textit{quadratically} with output length~\cite{tang2024questqueryawaresparsityefficient}.
For example, when serving Qwen3-8B~\cite{qwenlmQwQ32BEmbracing} on an H100 with a batch size of $128$ and an output of $8192$, loading the \kvc{} takes on average $21$ ms per step, accounting for over $70$\% of the end-to-end latency.

\begin{figure*}[!t]
    \centering
    \includegraphics[width=\textwidth]{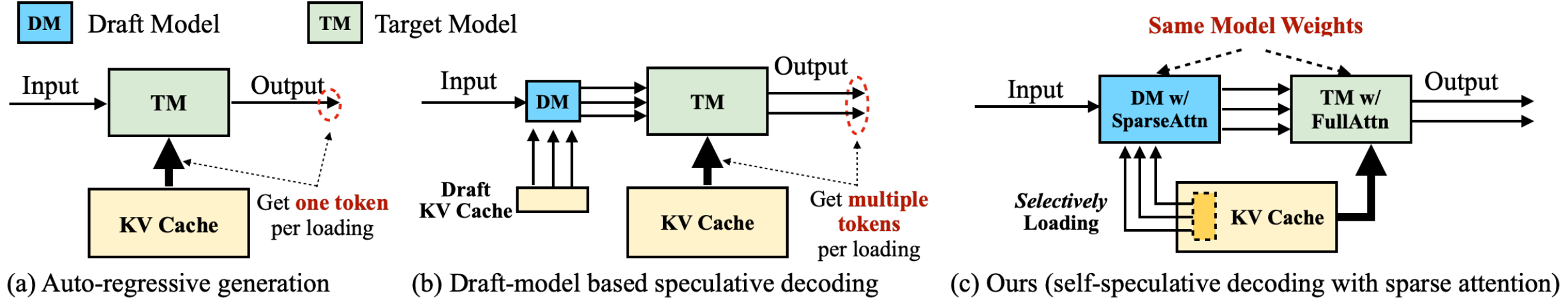}
    \caption{Comparison between autoregressive generation (a), draft model-based speculative decoding (b), and \sys{}.
\sys{} identifies the \kvc{} loading as the key bottleneck during token generation, and uses the same model weights with dynamic sparse attention as the draft model to achieve high efficiency and a high acceptance rate without additional training.
} 
    \label{fig:teaser}
\end{figure*}

To mitigate the memory bandwidth bottleneck, researchers have proposed a lossless technique, \textit{speculative decoding}~\cite{chen2023acceleratinglargelanguagemodel}.
In a nutshell, speculative decoding employs a smaller and faster \textit{draft model} to generate multiple candidate tokens sequentially. 
These candidates are then verified in parallel by the original \textit{target model}. 
This requires reading the large \kvc{} \textit{only once}. 
In contrast, without speculation, the entire \kvc{} needs to be loaded for every token. As a result, speculative decoding substantially reduces memory access and improves throughput.

However, existing speculative decoding methods require additional training or modification for each target model, limiting their applicability. 
Specifically, some solutions require training a separate standalone draft model~\cite{chen2023acceleratinglargelanguagemodel}; others modify the model architecture by adding decoder layers~\cite{li2025eagle3scalinginferenceacceleration}. 
These additional steps increase the complexity of real-world deployment. 
For example, training a small draft model requires careful data curation for a specific workload and may not generalize~\cite{liuosd2024onlinespeculativedecoding}. 
Moreover, deployment also needs to redesign the inference framework to orchestrate both models efficiently~\cite{Miao_2024}. 
Ultimately, this creates significant barriers to adoption~\cite{Zhang_2024}.

Other work explores training-free methods, which either employ rule-based heuristics (e.g., N-Gram~\cite{fu2024breaksequentialdependencyllm}) or leverages the model itself via self-speculation~\cite{liu2024speculativedecodingearlyexitingfaster,sun2024triforce}. 
For example, MagicDec~\cite{chen2024magicdec} adopts the full model with a \textit{sliding window attention} to draft tokens, followed by a full attention to verify. 
As the sparse attention greatly reduces the memory accessby up to $95$\% of the full attention, the overall throughput is improved correspondingly.
However, existing methods fail to provide ideal speedups for RLMs, due to its unique \textit{algorithmic} and \textit{systemetic} challenges (\refsec{sec:motivation:existing}).

Algorithmically, existing methods provide inaccurate drafted tokens, due to the lack of adaptivity to the high \textit{context dynamics} in RLMs. 
These reasoning models are inherently trained to generate diverse contexts~\cite{deepseekai2025deepseekr1incentivizingreasoningcapability,kimiteam2025kimik15scalingreinforcement}. 
For example, when solving a math problem, the model explores alternate solutions, resulting in dynamic contexts~\cite{guan2025rstarmathsmallllmsmaster}. 
Systemetically, speculative decoding and RLMs jointly introduce the following unique challenges: 
(1) \textit{workload flunctuation}: as the draft and verify phases have heterogeneous resource usages, the workload is imbalanced across iterations, leading to hardware underutilization;
(2) \textit{explicit synchronization}: the draft and verify phases require synchronization to ensure the validation of the drafted tokens, which prevents the expensive CPU operations from being overlapped with the GPU operations;
(3) \textit{\kvc{} underutilization}: the unpredictability of the output lengths from RLMs makes it difficult to fully saturate the \kvc{}, leading to GPU memory waste.

To alleviate these challenges, we present \sys{}, a \textit{lossless} and \textit{training-free} acceleration framework specialized for RLMs inference. 
In a nutshell, \sys{} reuses the same (target) model as the draft model, with a novel \textit{dynamic sparse attention} as the drafting mechansim. 
Co-designed with several system innovations, \sys{} fully unleashes the potential of self-speculation for RLMs.

At the core of \sys{} lies \textit{\alg{}}, a dynamic sparse attention tailored for speculative decoding on RLMs (\refsec{sec:design:sparse-attention}). 
\alg{} selectively loads and computes only the \textit{critical tokens}---those with the highest attention scores during inference—thereby significantly reducing memory bandwidth usage.
To handle context dynamics in reasoning, \alg{} identifies critical tokens by leveraging the exact attention scores obtained by full attention in each verification phase. 
These critical tokens are then used in the following draft steps, enabling high speculation accuracy through sparsity that adapts to various contexts~\cite{zhang2025pqcacheproductquantizationbasedkvcache,yang2024posttrainingsparseattentiondouble}.
With such co-design with speculative decoding, \alg{} achieves \textit{zero memory overhead} critical token identification compared to existing dynamic sparse methods.

In addition to \alg{}, \sys{} introduces innovations that address the three systemetic challenges above mentioned: 
(1) a \textit{unified batch scheduler} (\refsec{sec:design:batch-scheduler}) that batches the sparse draft and dense verification into a single batch to amortize weight loading, and evenly distributes requests across draft and verify so per-iteration resource usage stays balanced, thus mitigating workload fluctuation; 
(2) \textit{delayed verification} (\refsec{sec:design:async}) that delays verification by one iteration to remove CPU verification from the critical path. 
After launching a unified batch at the ($i$-1)-th iteration, the CPU asynchronously prepares all metadata for the $i$-th iteration excluding verification requests, while verification requests are postponed to the ($i$+1)-th iteration to overlap CPU work from ($i$-1)-th verification with GPU operations at $i$;
(3) a \textit{dynamic \kvc{} manager} (\refsec{sec:design:offloading}) that offloads/loads to and from CPU memory via an asynchronous and chunk-wise manner, thereby maximizing \kvc{} utilization.

We prototype and evaluate \sys{} across various reasoning models, including Qwen3-1.7B/8B/14B~\cite{qwenlmQwQ32BEmbracing}, using NVIDIA DGX-H100 servers with various tensor parallelism configurations. 
On real-world workloads such as AIME~\cite{huggingfaceAIMOaimovalidationaimeDatasets}, OlympiadBench~\cite{he2024olympiadbenchchallengingbenchmarkpromoting}, and LiveCodeBench~\cite{jain2024livecodebenchholisticcontaminationfree}, \sys{} achieves up to \uptocomparedtovllm{} throughput improvement compared to the \sota{} serving framework vLLM~\cite{kwon2023efficientmemorymanagementlarge}. 
Furthermore, when compared to existing training-free methods vLLM-NGram, MagicDec~\cite{chen2024magicdec}, and TriForce~\cite{sun2024triforce}, \sys{} achieves up to $1.56\times$, $1.36\times$ and $1.76\times$ throughput gain, respectively.

Our paper provides the following contributions:
\begin{itemize}[nosep]
    \item We analyze the performance bottleneck of batch RLMs inference, and pinpoint the potential benefit of sparse self-speculation with a theoretical formulation.
    \item We prototype a lossless and training-free acceleration framework, \sys{}, with a novel sparse attention \alg{},
    a unified batch scheduler, delayed verification, and a dynamic \kvc{} manager.
    \item We comprehensively evaluate our framework with real-world workloads, demonstrating \uptocomparedtovllm{} speedup compared to \sota{} inference frameworks.
\end{itemize}

\section{Background}
\label{sec:background}

\subsection{Transformer-based Large Language Models}
\label{sec:bg:llm}

Transformer-based LLMs mainly consist of two components: a multi-layer perceptron (MLP) and a self-attention module.
However, MLP and attention exhibit different performance characteristics during batch inference: MLP is typically compute-bound, while attention is memory-bound~\cite{tang2024questqueryawaresparsityefficient}.
For general matrix-matrix multiplication (GEMM) operations in MLP, batched requests share the same model weight, which can amortize memory loading cost, resulting in high arithmetic intensity. 
Therefore, considering a latency function $T_{\text{GEMM}}(B)$ that maps the batch size $B$ to the latency, it remains nearly constant before a certain threshold $\hat{B}$, and then increases linearly with $B$ after GPU is fully saturated~\cite{zhao2024atomlowbitquantizationefficient}.
In contrast, different requests have independent context (i.e., \kvc{}), which cannot be amortized across a batch. 
Thus, memory traffic scales linearly with sequence length and batch sizes, yielding consistently low arithmetic intensity and a linear latency function $T_{\text{Attn}}(M)$, where $M$ is the total memory size of \kvc{}.
In this paper, we focus on optimizing memory-bound attention, which is the primary bottleneck for long-output RLMs inference (\refsec{sec:motivation:mem-bound}).

\subsection{Evolving Workloads of RLMs}
\label{sec:bg:reasoning}

Reasoning language models (RLMs) are a class of LLMs designed to solve complex tasks through enhanced reasoning capabilities, as demonstrated by OpenAI-o1~\cite{openai2024openaio1card} and DeepSeek-R1~\cite{deepseekai2025deepseekr1incentivizingreasoningcapability}.
Unlike traditional LLMs, they are explicitly incentivized to follow deliberate CoT via reinforcement learning (RL)~\cite{kimiteam2025kimik15scalingreinforcement}.
Therefore, RLMs typically generate significantly more tokens during inference than non-reasoning models.  
To demonstrate this, we collect the average input and output lengths of several reasoning tasks. As shown in~\reftable{tab:avg_lengths}, the RLM Qwen3-14B generates on average $13542$ tokens on the AIME dataset, approaching $7\times$ more tokens than only $2593$ tokens from a non-reasoning Qwen-2.5-32B.

Such lengthy outputs pose significant challenges for both the \textit{training} and \textit{deployment} phases of RLMs. 
During RL post-training, RLMs are deployed to generate multiple \textit{rollouts}, which are then scored by the reward model to evolve the model. 
Such rollouts generation is throughput-oriented offline inference, which could take more than $90$\% of the end-to-end RL training time~\cite{deepscaler2025}. 
Besides, the deployment of RLMs including chatbots and agentic workflows is latency-oriented online inference, which is also bottlenecked by the memory-bound attention. 
In this paper, we focus on lossless acceleration that can be applied to both training and deployment phases.

\begin{table}[t]
  \centering
  \footnotesize
  \resizebox{\columnwidth}{!}{
  \begin{tabular}{lccc}
    \toprule
    \multirow{2}{*}{\textbf{Dataset}}
      & \multirow{2}{*}{{\textbf{Avg.\ Input}}}
      & \multicolumn{2}{c}{\textbf{Avg.\ Output (mean ± std)}} \\
    \cmidrule(lr){3-4}
      &  & \textbf{Qwen3-14B} & \textbf{Qwen2.5-32B-Instruct} \\
    \midrule
    \textbf{AIME}           & \textit{138} & \textit{13185 $\pm$ 7626} & \textit{1732 $\pm$ 997} \\
    \textbf{OlympiadBench}   & \textit{124} & \textit{10233 $\pm$ 7889} & \textit{957 $\pm$ 728} \\
    \textbf{LiveCodeBench}  & \textit{148} & \textit{10254 $\pm$ 7458} & \textit{618 $\pm$ 157} \\
    \bottomrule
  \end{tabular}
  }
  \caption{Average (mean ± std) token lengths on three reasoning evaluation datasets.  
           Input length is identical for both models; output lengths include
           their standard deviations.}
  \label{tab:avg_lengths}
\end{table}

\subsection{Speculative Decoding}
\label{sec:bg:speculative}
To mitigate the memory-inefficiency, speculative decoding is proposed as a \textit{lossless} acceleration by trading off computation for memory~\cite{chen2023acceleratinglargelanguagemodel, fu2024breaksequentialdependencyllm}. 
The core idea is to leverage a lightweight draft model to propose $k$ candidate tokens sequentially, which are then verified in parallel by the original target model.
Candiate tokens that align with the target model are accepted, while the rest are discarded and regenerated via the same workflow, ensuring lossless generation quality of the original model.

Assuming the latency of draft and target models are $T_{\text{draft}}$ and $T_{\text{target}}$, respectively, the acceleration comes from the fact that $T_{\text{draft}} \ll T_{\text{target}}$. 
Therefore, despite only $\alpha\cdot k + 1$ candidate tokens are accepted (with an acceptance rate $\alpha$ and a bonus token), the end-to-end latency when enabling speculative decoding is $(k\cdot T_{\text{draft}} + T_{\text{verify}})$ is smaller than the latency $(\alpha k + 1)\cdot T_{\text{target}}$ without speculative decoding.

\section{Motivation}
\label{sec:motivation}

\subsection{Batch RLMs Inference is Attention-bound}
\label{sec:motivation:mem-bound}
The long-generation of RLMs introduces a significant memory bottleneck in batched inference. 
Unlike short-generation tasks, each reasoning request accumulates a large \kvc{} until completion, rapidly exhausting GPU memory and severely constraining the number of concurrent requests (i.e., batch size).
For example, serving \lm{}-3-8B on an H100 can accommodate only $64$ requests with an $8$K generation length. 
Inference at such small batch sizes is memory-bound for both MLP and attention, and with the \kvc{} often larger than model weights, attention emerges as the dominant runtime component.

\begin{figure}[!t]
    \centering
    \includegraphics[width=0.9\linewidth]{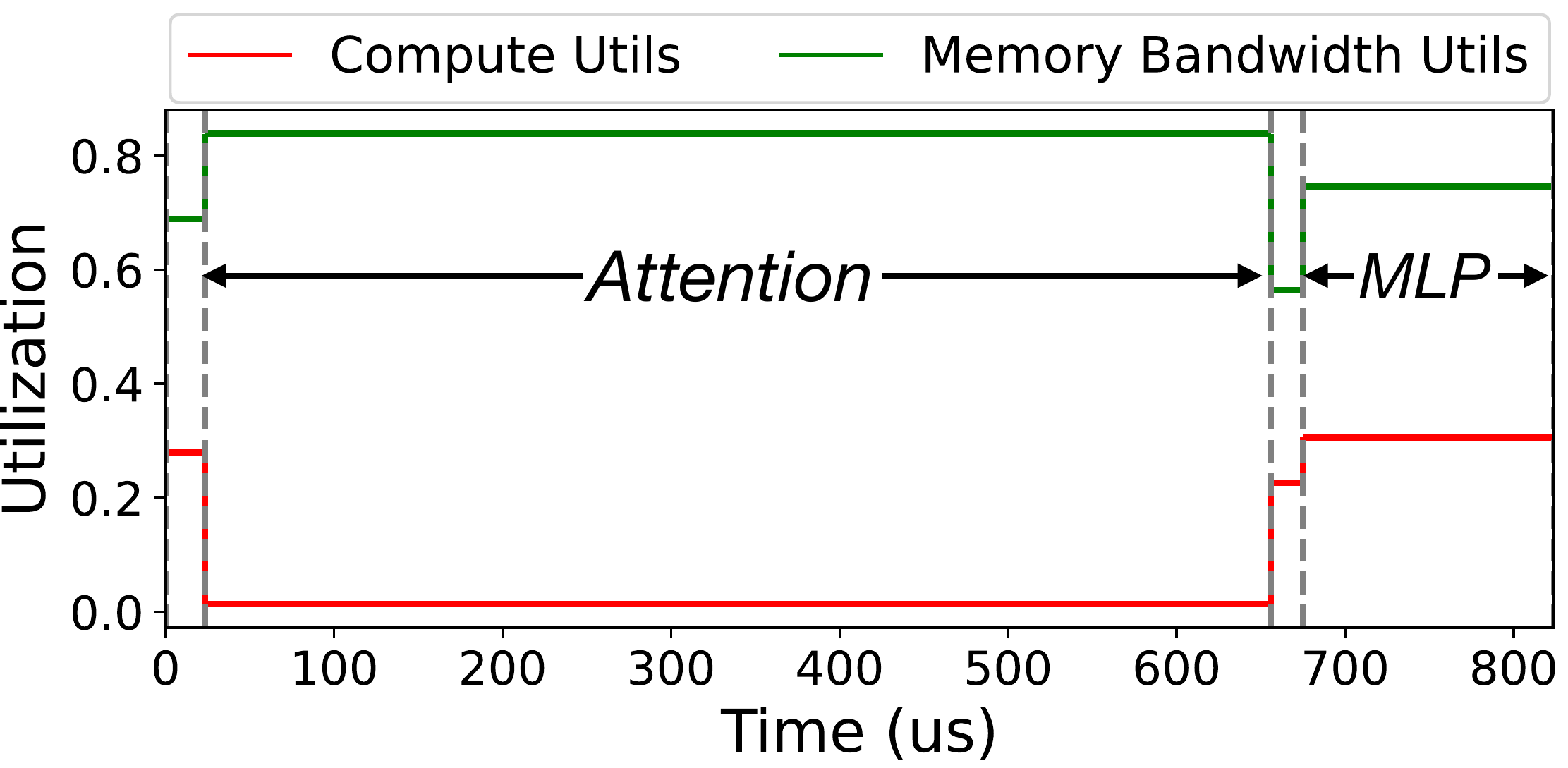}
    \caption{Compute and memory bandwidth utilization of Qwen3-8B on an H100, with an average output of $12$K on AIME. }
    \label{fig-motivation-utils}
\end{figure}

To demonstrate this, we profile bandwidth and compute utilization over end-to-end execution for RLMs running on vLLM. 
We visualize the utilization within a single iteration in~\fig{fig-motivation-utils}. 
Compute is consistently underutilized below $50$\% even when executing MLP; 
In contrast, memory bandwidth is heavily used across the entire iteration, indicating a memory-bound workload. 
Although both MLP and attention are memory-bound, attention is dominant and accounts for more than $77$\% of end-to-end execution time.

\subsection{Sparse Self-Speculative Decoding Helps}
\label{sec:motivation:sspec}

\MyPara{Self-speculative decoding with sparse attention.} 
As discussed in~\refsec{sec:motivation:mem-bound}, 
the RLMs inference is attention-bound due to the execessive memory access of \kvc{} under batch scenarios. 
Fortunately, prior studies on sparse attention show that a small subset of tokens in \kvc{} (e.g., $5$\%) dominates the attention output, providing nearly lossless speedup~\cite{tang2024questqueryawaresparsityefficient,lin2025twilight}.
Therefore, by selectively loading and computing the most critical tokens, memory access can be substantially reduced, trading off accuracy for efficiency. 
To ensure lossless acceleration, such methods can be applied as the draft model in speculative decoding, working with the full target model (i.e. sparse self-speculative decoding~\cite{ sun2024triforce}).

\MyPara{Theoretical analysis of speedup.}
We now formulate the \textit{theoretical speedup} of sparse self-speculative decoding, to estimate the potential improvement. 
We define the speedup $\eta$ as the ratio of the average generation latency per token with and without speculative decoding (i.e., $T_{\text{base}} / T_{\text{spec}}$).
Specifically, let $M$ be the total memory size of \kvc{} and $B$ be the number of concurrent requests. 
With the same notations in~\refsec{sec:bg:llm}, $T_{\text{base}}=T_{\text{GEMM}}(B) + T_{\text{Attn}}(M)$, ignoring other small operations for simplicity.

Assuming $k$ the \dl{}, $\alpha$ the \ar{}, and $s$ the sparsity ratio of the draft attention, we calculate $T_{\text{spec}}$ by considering an entire round of speculation with $k$ draft and $1$ verification phases.
Thus, we have $(k \alpha + 1)\cdot T_{\text{spec}} = k\cdot T_{\text{draft}} + T_{\text{verify}}$, where $T_{\text{draft}} = T_{\text{GEMM}}(B) + T_{\text{Attn}}(M \cdot s)$ and $T_{\text{verify}} = T_{\text{GEMM}}((k+1)\cdot B) + T_{\text{Attn}}(M)$. 
As the $B$ concurrent requests are randomly distributed across the $k$ draft and $1$ verification stages, the average input size for GEMMs is $\frac{2k + 1}{k+1}B$, while the average input size for attention is $\frac{ks + 1}{k+1}M$. 
Thus, we can simpify the previously derived equation of $(k \alpha + 1)\cdot T_{\text{spec}}$ as:
\[
(k+1)(T_{\text{GEMM}}(\frac{2k + 1}{k+1}B) + T_{\text{Attn}}(\frac{ks + 1}{k+1}M))
\]
As discussed in~\refsec{sec:bg:llm}, $T_{\text{GEMM}}$ is non-linear function when GPU is not fully saturated, while $T_{\text{Attn}}$ is approximately linear. 
Thus, the $T_{\text{spec}}$ can be approximated as:
\[
\frac{k+1}{k\alpha+1} T_{\text{GEMM}}(\frac{2k + 1}{k+1}B) + \frac{ks+1}{k\alpha+1}T_{\text{Attn}}(M)
\]
Given the formulation of latency per accepted token $T_{\text{spec}}$, and the latency per token $T_{\text{base}}$ on vanilla generation, the speedup $\eta$ is derived as $\eta=T_{\text{base}} / T_{\text{spec}}$.

\MyPara{Pratical performance implications.}
As GEMM and attention have dramatically different characteristics, we analyze the implications of $\eta$ over GEMM and attention separately.

For attention, since $T_{\text{Attn}}(M)$ dominates $T_{\text{base}}$ in RLMs inference as discussed in~\refsec{sec:motivation:mem-bound}, the reduction ratio of \kvc{} access is $(ks+1)/(k\alpha+1)$, contributing the most to the speedup.
Since $\alpha \gg s$ in practice, the attention latency can be greatly reduced. 
For example, with $k=16$, $\alpha=0.75$, $s=0.05$, the attention latency is reduced by $6.78\times$.

However, such memory reduction comes with the cost of extra GEMM computations, trading off computation for memory.
In practice, to avoid untolerable latency increase, the batch size should typically be limited to a certain threshold $\hat{B}$, so that $T_{\text{GEMM}}$ is not oversaturated.
For example, on Hopper GPUs, $\hat{B}=256$ only incurs minimal latency increase. 
Besides, $\alpha$ is also crucial to limit the latency increase, as the speedup $\eta$ is inversely proportional to $\alpha$.

\subsection{Existing Methods Fall Short on RLMs}
\label{sec:motivation:existing}

\begin{figure}[!t]
    \centering
    \vspace{4.5mm}
    \includegraphics[width=0.95\linewidth]{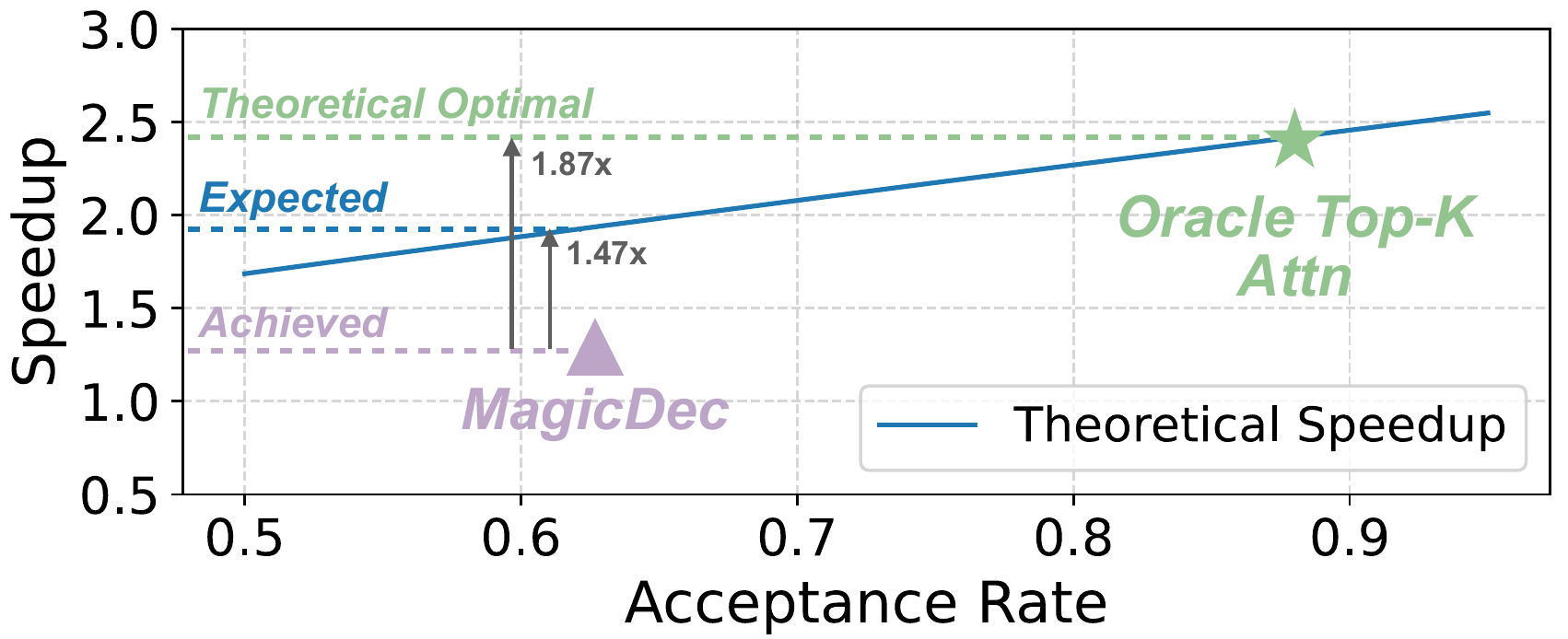}
    \caption{Theoretical and achieved speedup over vLLM of MagicDec and self-speculation with oracle Top-K attention. Assume a sparsity ratio $s=0.5$ and speculative step $k=8$.}
    \label{fig-motivation-speedup}
\end{figure}

\begin{figure}[!t]
    \centering
    \includegraphics[width=0.85\linewidth]{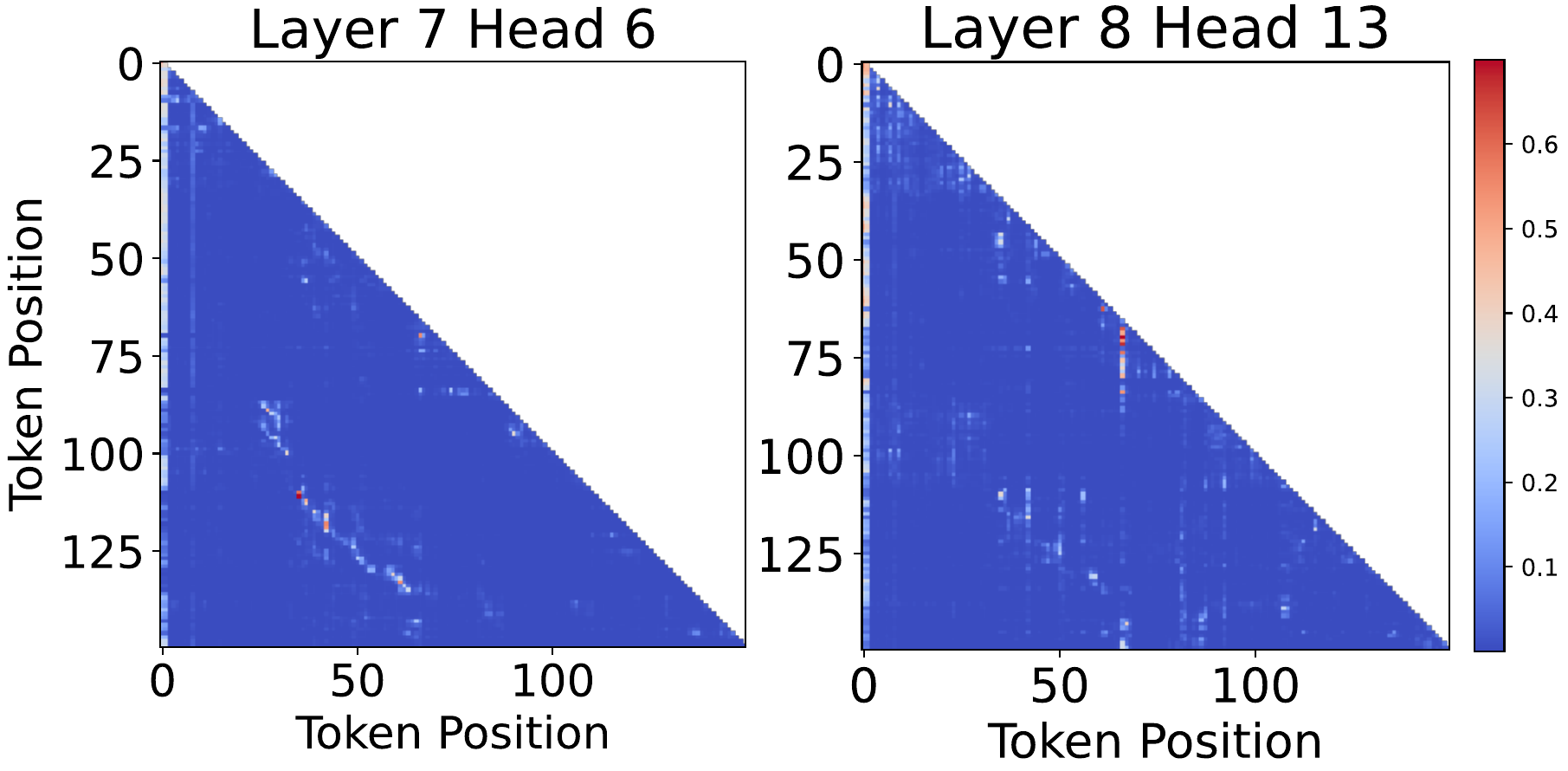}
    \caption{Visualization of attention scores of Qwen3-8B on the AIME dataset. While the attention pattern has spatial locality, it undergoes substantial changes during generation.}
    \label{fig-design-attn}
\end{figure}

However, we find that existing methods fail to provide expected speedup under RLMs inference, as modeled in~\refsec{sec:motivation:sspec}. 
As shown in~\fig{fig-motivation-speedup}, when serving Qwen3-8B on an H100 with AIME, MagicDec fails to approach the acceptance rate of the \textit{oracle top-k attention}~\cite{lin2025twilight}, thus falling far behind the theoretical optimal speedup. 
This is because the draft model in existing methods is not adaptive to the \textit{context dynamics} in RLMs, leading to inaccurately drafted tokens.
Furthermore, even with the same acceptance rate, existing methods still have a non-negligible gap between the expected speedup, due to several \textit{system challenges}.

\MyPara{Context dynamics.}
For long output of RLMs, sparsity patterns shift significantly based on semantic context during generation. 
As shown in~\fig{fig-design-attn}, the critical tokens of the \kvc{} differs dramatically over time. 
Therefore, instead of using a static sparsity pattern (e.g., sliding window attention), a dynamic sparse attention mechansim that adapts to the context dynamics is necessary to achieve a high acceptance rate $\alpha$ during drafting. 
Moreover, such sparse attention design should incur minimal computation or storage overhead, to avoid adding complexity to the system.

\MyPara{Workload fluctuation.}
As draft and verify phases have heterogeneous resource usages, naive scheduling these two phases separately leads to workload fluctuation across iterations, incurring substantial hardware underutilization. 
This heterogeneity mainly comes from GEMM operations. 
For instance, given a batch size $B$ and speculative steps $k$, a sequential execution pattern (all draft phases followed by one verification phase) requires: 
(1) $k$ GEMM operations with input size $B$, and 
(2) one verification GEMM with input size $(k+1)B$.
This leads to under-utilization during drafting and oversaturation during verification, with a total time of $kT_{\text{GEMM}}(B) + T_{\text{GEMM}}((k+1)B)$. 
Instead, a uniform scheduling incurs a total time of $(k+1)T_{\text{GEMM}}(\frac{2k + 1}{k+1}B)$. 
Considering the non-linearity of $T_{\text{GEMM}}$ before saturation, $T_{\text{GEMM}}(\frac{2k + 1}{k+1}B)< T_{
\text{GEMM}}(2B)\approx T_{\text{GEMM}}(B)$ when $B$ is far below the saturation point. 
Therefore, the naive scheduling incurs up to $2\times$ longer execution time.

\MyPara{Explicit synchronization.}
The paradigm of speculative decoding, i.e., drafting candidate tokens followed by verification, introduces an explicit synchronization between CPU and GPU, preventing the expensive CPU operations from being overlapped with the GPU operations. 
Specifically, the GPU must wait for the CPU to complete the verification process, which cleans metadata of rejected tokens from reqeusts and resets draft status for the next generation step. 
As a result, \sota{} inference frameworks~\cite{zheng2024sglangefficientexecutionstructured,kwon2023efficientmemorymanagementlarge} must run CPU operations on the critical path when enabling speculative decoding, incurring significant overhead. 
Such inefficiency is further exacerbated in \sspec{}, where each iteration has verification requests under unified scheduling.

\begin{figure}[!t]
    \centering
    \includegraphics[width=0.9\linewidth]{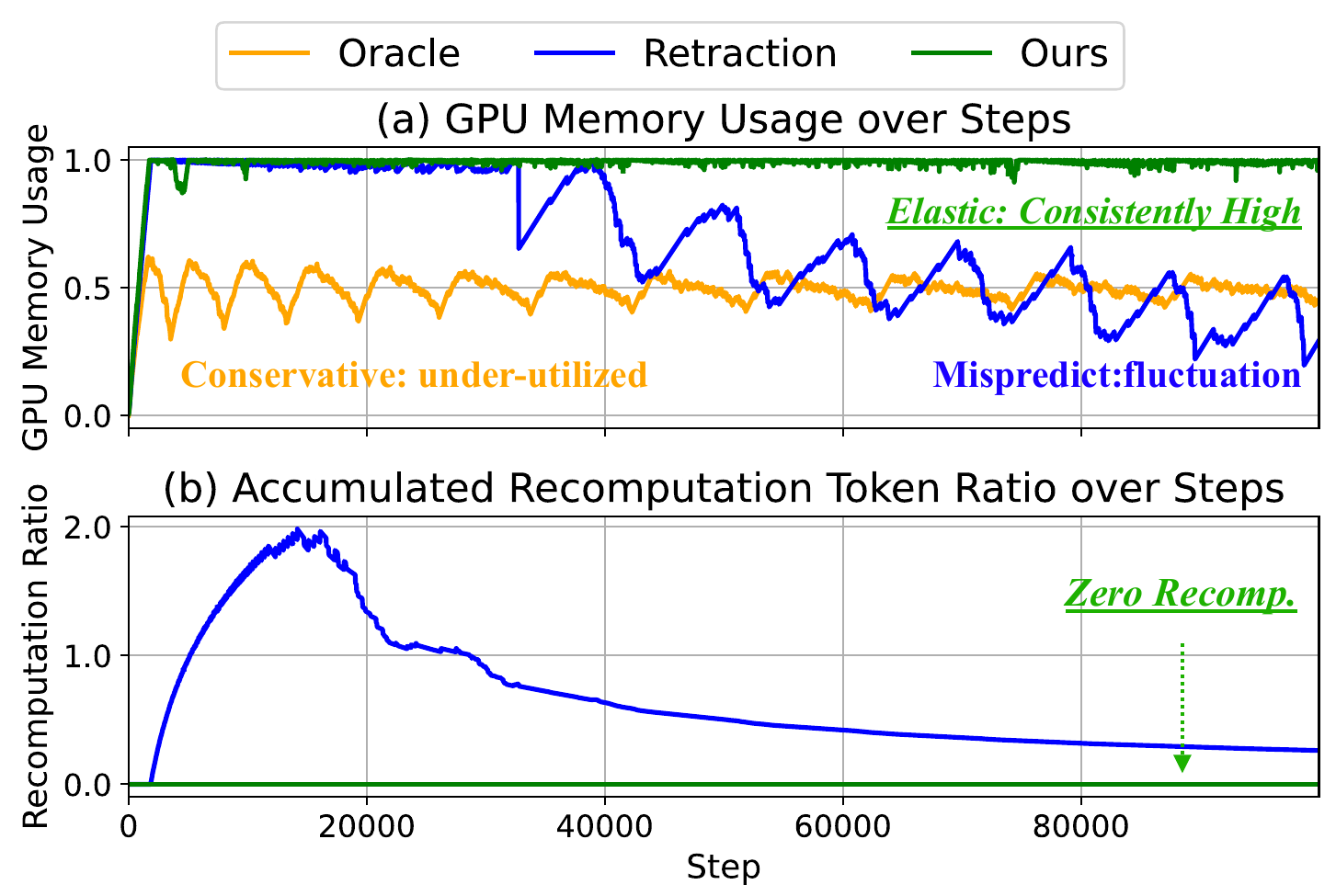}
    \caption{The memory utilization and recomputation ratio during the first $100$K steps when serving Qwen3-8B with AIME on H100. Existing methods either underutilizes \kvc{} capacity, or leads to excessive recomputation. Our \dkvm{} can fully utilize the capacity without incurring recomputation.}
    \label{fig-design-offloading}
    \vspace{-4mm}
\end{figure}

\MyPara{\kvc{} underutilization.}
The huge variance of output lengths from RLMs (\refsec{sec:bg:reasoning}) makes it hard to perfectly manage \kvc{}, as the output lengths are unknown before generation. 
We collect the memory utilization and recomputation (i.e., the ratio of recomputed tokens to generated unique tokens) from a real-world trial. 
As shown in~\fig{fig-design-offloading}, existing methods either underutilize \kvc{} capacity or lead to excessive recomputation due to misprediction. 
Such underutilization of \kvc{} 
severely limits the self-speculation speedup, as the efficiency metric $\eta$ decreases with smaller \kvc{} size ({\refsec{sec:motivation:sspec}}).

\subsection{\sys{}: Unleashing Sparse Self-Speculation with Efficient Algorithm-System Co-design}
\MyPara{Opportunities.}
In this paper, we aim to leverage the sparse self-speculative decoding to mitigate the memory bandwidth bottleneck of long-generation from batch RLMs inference, providing lossless and training-free acceleration. 

\MyPara{Challenges.}
However, realizing the theoretical speedup of self-speculation presents several challenges, requiring algorithm-system co-design:
(1) how to design a sparse attention that adapts to the context dynamics with minimal overhead;
(2) how to schedule draft and verify phases to avoid workload fluctuation;
(3) how to handle the explicit synchronization to enable CPU/GPU overlap; and 
(4) how to fully utilize the \kvc{} capacity without recomputation.

\section{Design}
\label{sec:design}
\begin{figure}[!t]
    \centering
    \includegraphics[width=\linewidth]{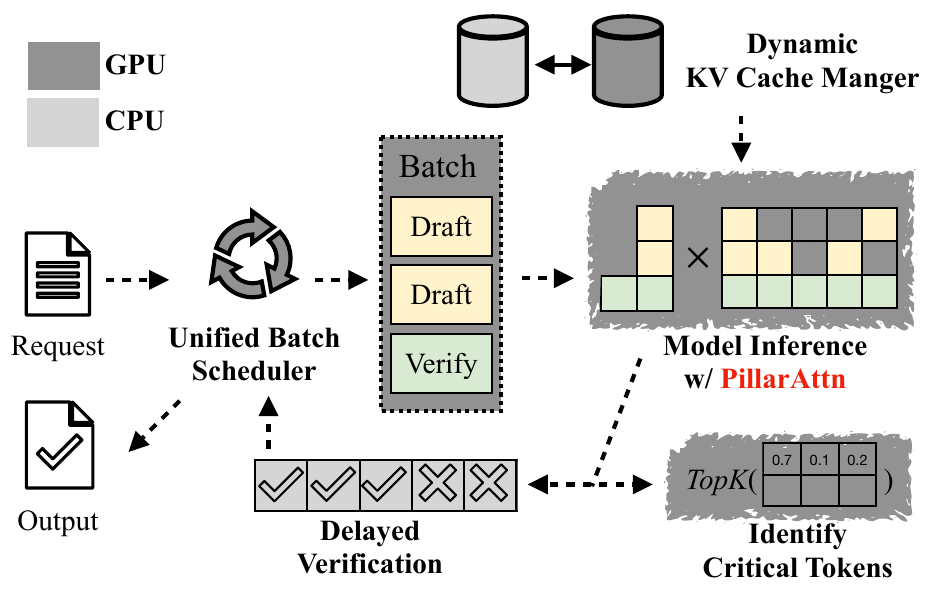}
    \vspace{-4mm}
    \caption{Overview of \sys{}.
} 
    \label{fig:sys}
    \vspace{-4mm}
\end{figure}

In this section, we describe the design of \sys{}'s components, including a dynamic sparse attention algorithm (\refsec{sec:design:sparse-attention}), a unified batch scheduler (\refsec{sec:design:batch-scheduler}), delayed verification (\refsec{sec:design:async}), and a dynamic \kvc{} manager (\refsec{sec:design:offloading}).
We visualize the design overview of \sys{} in \fig{fig:sys}.

\subsection{\alg{}: Dynamic Sparse Attention Tailored for Self-Speculative Decoding}
\label{sec:design:sparse-attention}

As discussed in~\refsec{sec:motivation:sspec}, both a high sparsity ratio $s$ and a high acceptance rate $\alpha$ are critical to the performance of sparse self-speculation. 
Therefore, a sparse attention that \textit{accurately} and \textit{efficiently} identifies sparsity patterns is necessary. 
To this end, \sys{} proposes \alg{}, a sparse attention mechanism tailored for self-speculation decoding. 

\MyPara{Dynamic sparsity pattern.}
To adapt to the context dynamics as discussed in~\refsec{sec:motivation:existing}, instead of using a fixed sparsity pattern (e.g., sliding window or static patterns from prompts), \alg{} periodically re-identifies and updates the sparsity pattern used in the drafting phase. 
Built on the assumption that the contextual semantics have spatial locality, \alg{} updates the sparsity pattern at a small stride, within which the sparsity patterns are fixed. 
Therefore, the overhead of identification can be amortized over a stride of iterations.

\begin{figure}[!t]
    \centering
    \includegraphics[width=1\linewidth]{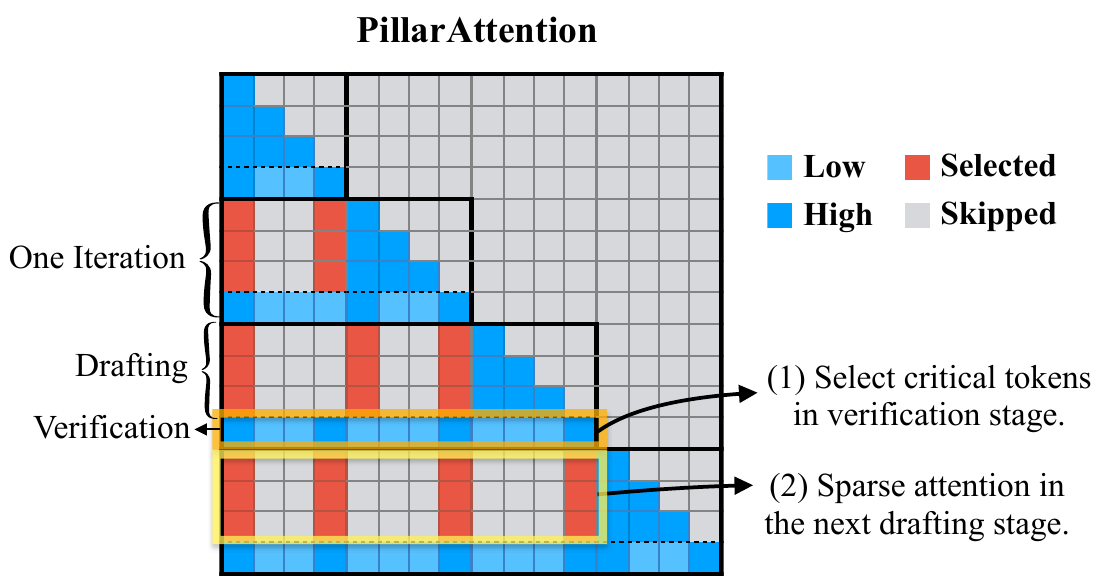}
    \vspace{-4mm}
    \caption{Illustration of \alg{}. \alg{} performs full attention and uses attention scores in the verification phase to identify sparsity patterns for the next $k$ draft phases.}
    \label{fig:pillarattn}
    \vspace{-4mm}
\end{figure}

\MyPara{Overhead-free identification.}
Such a stride is co-designed with the paradigm of self-speculative decoding, with the same value as the speculative steps $k$. 
After every $k$ draft steps with sparse attention, a verification step with full attention is performed, which computes the attention scores for all tokens in the \kvc{} as intermediate results. 
To avoid extra computation and storage overhead, \alg{} on-the-fly dumps the attention scores during verification via customized attention kernels. 
Therefore, the sparsity pattern can be directly identified by reusing and applying Top-K on the dumped attention scores, which are first averaged over $k$ draft tokens and query heads (within same group) if group query attention is applied. 
Specifically, the attention logits and logarithm summation of exponential are cached during verification, which are used to rematerialized attention scores for identification. 
We visualize the workflow of \alg{} in~\fig{fig:pillarattn}, where the sparsity pattern is updated during verification after every $3$ draft phases.

\subsection{Unified Batch Scheduler}
\label{sec:design:batch-scheduler}

As discussed in~\refsec{sec:motivation:existing}, the draft and verification phases have heterogeneous resource usages, which leads to hardware underutilization thus suboptimal performance. 
To address this, \sys{} introduces a unified batch scheduler, that provides a uniform abstraction and a workload-aware scheduling for both phases. 
Furthermore, to enhance hardware efficiency, \sys{} introduces a fused attention kernel that effciently batch sparse and full attention into a single kernel.

\MyPara{Uniform abstraction for draft and verification.}
Since draft and target models share the same model weights in self-speculation, both phases go through the exactly same on-GPU data and control flow, except for sparse attention and full attention. 
Fortunately, modern inference frameworks already integrate PagedAttention~\cite{kwon2023efficientmemorymanagementlarge}, which is essentially a sparse attention implementation at the granularity of the page. 
Therefore, by using a page size of $1$, both sparse and full attention can be unified with the same pipeline, which greatly simplifies the system design and unleashes the flexibility of scheduling in both phases.

\begin{figure}[!t]
    \centering
    \includegraphics[width=0.95\linewidth]{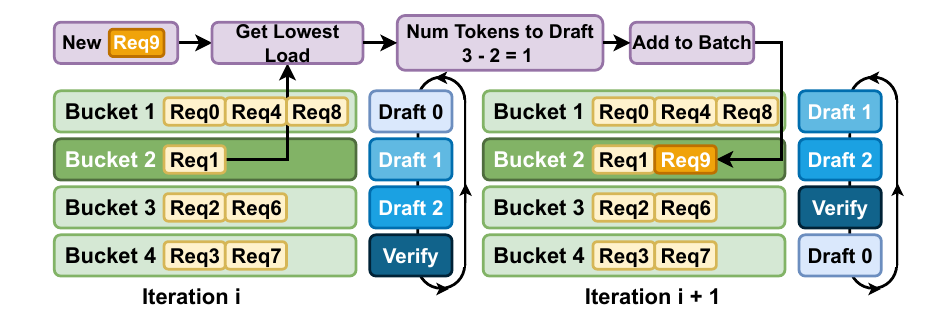}
    \caption{Illustration of scheduling new requests.
\UBS{} identifies the least-loaded bucket and schedules new requests to it by adjusting the \dl{}.
    }
    \label{fig:batch-scheduler}
    \vspace{-4mm}
\end{figure}

\MyPara{Workload-aware scheduling.}
As shown in~\refsec{sec:motivation:existing}, sequential execution of all draft phases followed by one verification phase results in poor hardware utilization, due to the extremely unbalanced workload. 
Therefore, given the scheduling flexibility enabled by the unified abstraction, \sys{} evenly mixes requests from both phases in each batch (at each generation step). 
To achieve perfect balance, \sys{} introduces a greedy bin-packing strategy that assigns incoming new requests into different draft phases in a workload-aware way. 
Specifically, \sys{} maintains $k$ buckets to track request counts for each draft phase and assigns each incoming request to the least-loaded bucket. 
This scheduling strategy is illustrated in \fig{fig:batch-scheduler}.

\MyPara{Fused sparse and full attention.}
Even under ideal scheduling, the attention operation still suffers from low hardware utilization due to its heterogeneity in the draft and verification phases. 
This is because the attention in draft and verification has different arithmetic intensities due to different input token counts, leading to different best kernel implementations (e.g., tile sizes and MMA instruction). 
To demonstrate this, we profile \sota{} FlashInfer~\cite{ye2025flashinferefficientcustomizableattention} for both sparse and full attention. 
Results show that the kernel optimized for verification (full) attention achieves near $85$\% bandwidth but less than $50$\% for draft (sparse) attention; 
kernels optimized for sparse attention deliver over $80$\% bandwidth, but degrade to $50$\% on full attention. 
To address this, \sys{} introduces a fused attention kernel in persistent-kernel style, which on-chip dispatches verification and draft attention into their best template instead of launching two kernels, achieving the best of both. 
We evaluate the fused kernel in~\refsec{sec:eval:ablation}.

\subsection{Delayed Verification}
\label{sec:design:async}
As detailed in~\refsec{sec:motivation:existing}, the verification phase introduces an explicit synchronization between CPU and GPU, preventing the expensive CPU operations from being overlapped with the GPU operations. 
For example, the execution of $i$-th iteration depends on the verification results of ($i-1$)-th, specifically on rejected tokens and the updated sparsity pattern. 
Such sequential execution can account for over $20$\% of end-to-end latency~\cite{wuklabMLSysWukLab}.

\begin{figure}[!t]
    \centering
    \includegraphics[width=0.95\linewidth]{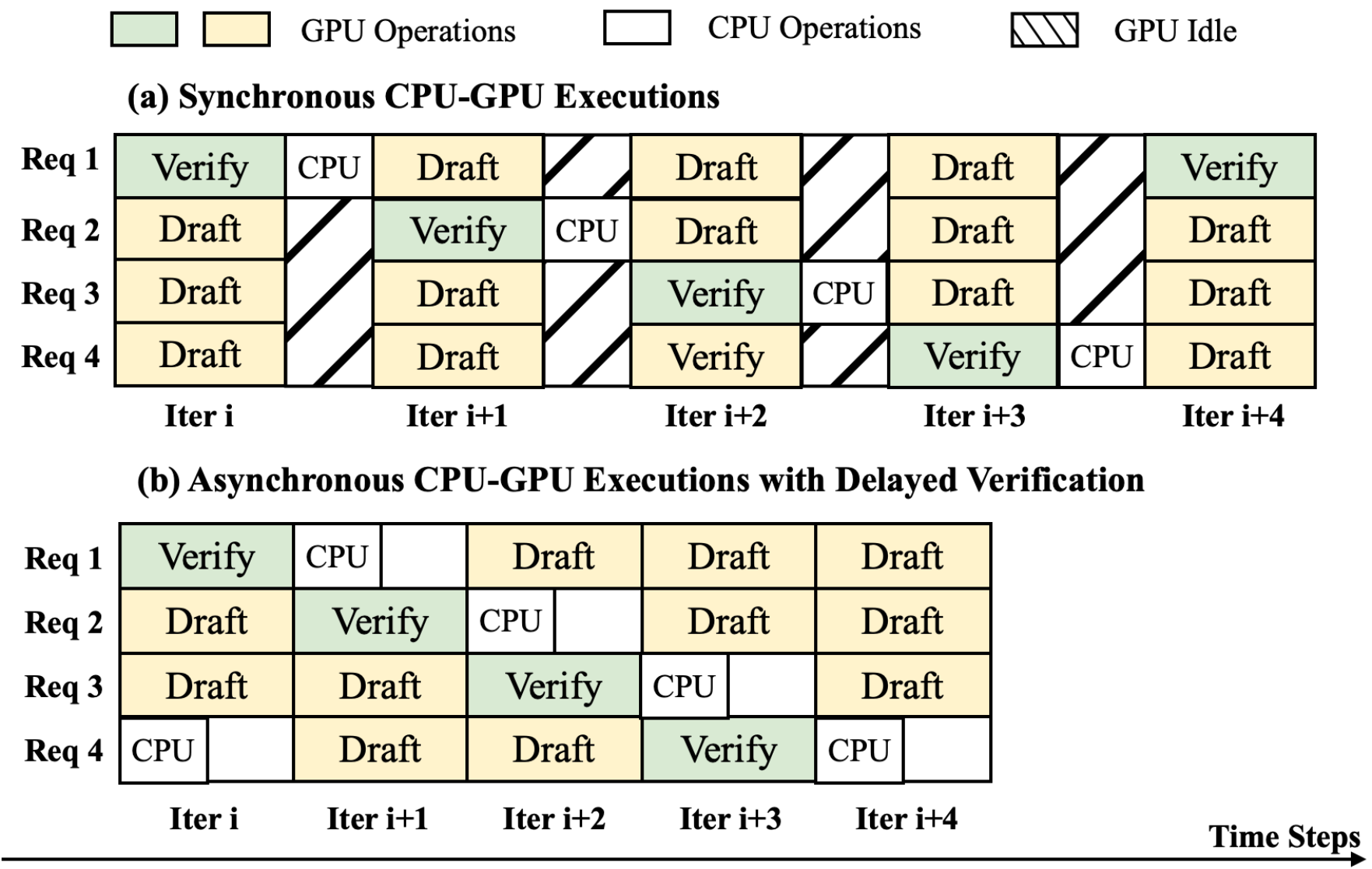}
    \caption{Illustration of delayed drafting to achieve asynchronous CPU-GPU execution.
Requests in the verification phase stall for one cycle for the CPU to determine the number of accepted tokens and the critical tokens, while other requests can still move forward.
}
    \vspace{-4mm}
    \label{fig:cpu-async}
\end{figure}

Our key observation is that such synchronization only applies to requests in the verification phase. 
Under balanced scheduling in~\refsec{sec:design:batch-scheduler}, such requests occupy only a small fraction ($\frac{1}{k+1}$) of the batch $B$. 
Therefore, instead of stalling the entire batch, \sys{} allows the metadata preparation for non-verification requests on CPU to proceed directly without waiting for the verification results from GPU. 
For example, only verification requests from ($i-1$)-th iteration are stalled and taken out from the $i$-th iteration. 
After getting the verification results from GPU, the requests are issued to the GPU for execution at ($i+1$)-th iteration.
We illustrate the delayed verification in~\fig{fig:cpu-async}.

\subsection{Dynamic \kvc{} Management}
\label{sec:design:offloading}

\MyPara{Aggressive CPU Offloading.} 
Given the high output length variance in RLMs (\reftable{tab:avg_lengths}), achieving high KV cache utilization without recomputation is challenging (\fig{fig-design-offloading}). 
Therefore, instead of optimizing output-length prediction, \sys{} prefers to aggressively increase request concurrency to fully utilize \kvc{}, while offloading \kvc{} to host once approaching out-of-memory to avoid recomputation. 
Note that both offloading and loading follow the FIFO order, assuring fairness and avoiding starvation.

\MyPara{Overhead analysis.} 
First, \sys{} achieves offloading with negligible overhead by breaking the large \kvc{} into chunks and offloading chunk-by-chunk asynchronously. 
For instance, when running Qwen3-8B on a single H100 GPU with a batch size of $128$, each decoding step generates just $128$ new tokens, requiring only $18$MB of \kvc{} memory~\footnote{$128*128*8*2*2*36/1e6=18$MB.}.
Since the GPU latency per iteration is on the magnitude of $10$ ms, the necessary bandwidth is only $18$ GB/s to overlap offloading with GPU computation, which is well below the PCIe bandwidth limit.
Besides, \sys{} prioritizes scheduling the offloaded requests whenever GPU has available memory. 
This strategy bounds worst-case CPU usage to GPU capacity, e.g., $640$GB for an $8\times$H100 server, well within typical CPU DRAM limits in data centers.

\section{Evaluation}
\label{sec:eval}

\begin{figure*}[!t]
    \centering
    \includegraphics[width=\textwidth]{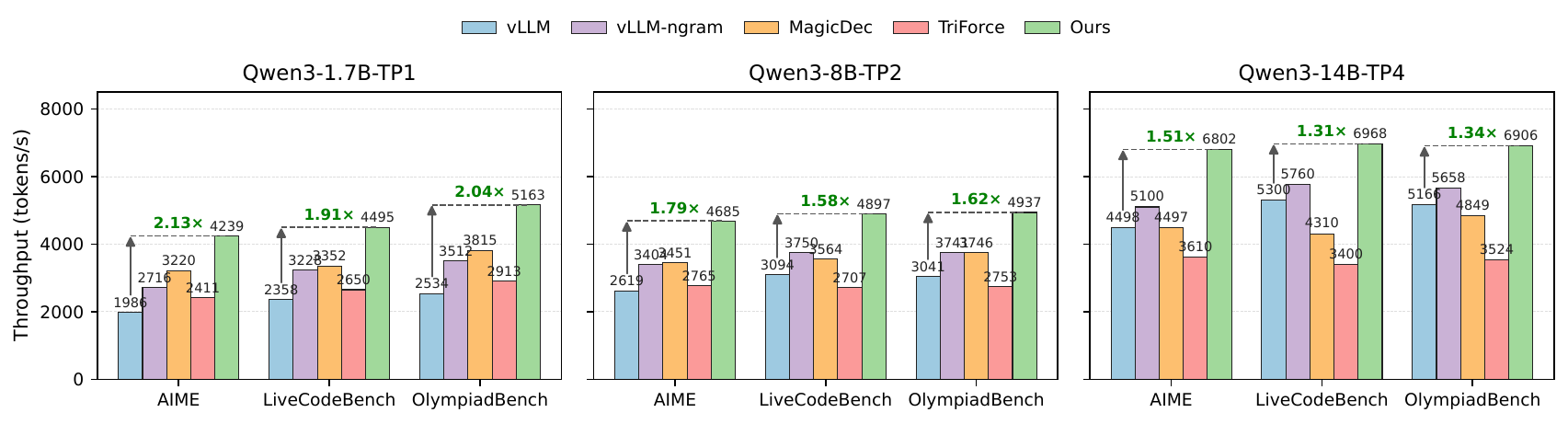}
    \vspace{-4mm}
    \caption{End-to-end throughput comparison of \sys{} and existing \textit{training-free} acceleration frameworks. } 
    \label{fig-e2e-tps}
    \vspace{-2mm}
\end{figure*}

\subsection{Evaluation Setup}
\label{sec:eval:setup}

\MyPara{Models and Hardwares.}
We evaluate \sys{} using three \sota{} open-sourced RLMs: (1) Qwen3-1.7B, (2) Qwen3-8B, and (3) Qwen3-14B, covering a wide range of model sizes and architectures. 
We use tensor parallelism (TP) to partition models across multiple GPUs, selecting the parallelization that achieves the highest throughput per GPU, with TP1/2/4 for 1.7B/8B/14B, respectively. 
We use NVIDIA DGX-H100-SXM5 GPUs as our testbed.

\MyPara{Datasets.}
We use the following datasets as target reasoning workloads spanning math, general knowledge, and programming: 
(1) \textit{AIME}~\cite{aime_1983_2024}: Math problems from the American Invitational Mathematics Examination; 
(2) \textit{OlympiadBench}~\cite{he2024olympiadbenchchallengingbenchmarkpromoting}: Olympiad-level bilingual scientific problems across STEM requiring advanced multi-step reasoning, and we use its text QA subset for evaluation; and 
(3) \textit{LiveCodeBench}~\cite{jain2024livecodebench}: Coding questions collected from LeetCode, AtCoder, and Codeforces.
For each workload, we randomly sample $2048$ requests to saturate the pipeline for the following evaluations. 
We set the temperature as $0.65$ following common practice. 

\MyPara{Baselines.}
We compare \sys{} against the following serving frameworks combined with different speculative decoding algorithms: 
(1) \textit{vLLM}\footnote{Release v0.11.0 with commit hash: b8b302c}: we enable vLLM-V1 by default for best performance; 
(2) \textit{vLLM-NGram}~\cite{leviathan2023fastinferencetransformersspeculative}: a training-free speculative decoding integreted into vLLM; 
(3) \textit{MagicDec}~\cite{chen2024magicdec}: as the original open-sourced one is hard-coded for short output lengths and cannot be easily adapted, we reproduced MagicDec within our framework stricly following their paper; and 
(4) \textit{TriForce}~\cite{sun2024triforce}: similar to MagicDec, we reproduced TriForce based on vLLM; and 
(5) \textit{vLLM-EAGLE3}~\cite{li2025eagle3scalinginferenceacceleration}: a \sota{} draft-model based speculative decoding algroithm integreted by vLLM. We adopt all EAGLE3 draft models open-sourced by Tencent~\cite{AngelSlim2025}. 
For all baselines, we enable chunked prefill, CUDA graph, and continuous batching. %
We set up the best configuration for both NGram ($k=4$) and EAGLE3 ($k=3$) according to our tests, with the tree-speculation disabled in batch inference following~\cite{li2025eagle3scalinginferenceacceleration}. 
We set the maximal batch size for all frameworks to $256$, which is large enough to saturate GPU memory in our setup.

\subsection{End-to-End Performance}
\label{sec:eval:overall}
\begin{figure}[!t]
    \vspace{6mm}
    \centering
    \includegraphics[width=0.85\columnwidth]{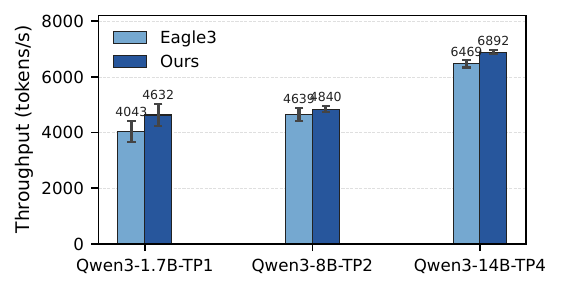}
    \vspace{4.5mm}
    \caption{End-to-end throughput comparison of \sys{} and draft model based methods, averaged over three datasets. Error bars denote the standard deviation across datasets.}
    \label{fig-eagle3-sspec}
\end{figure}

\MyPara{Ours vs. training-free speculative decoding.}
\sys{} is a training-free speculative decoding system by design. 
Thus, we compare \sys{} with training-free speculative decoding methods including vLLM-NGram~\cite{leviathan2023fastinferencetransformersspeculative}, MagicDec~\cite{chen2024magicdec}, and TriForce~\cite{sun2024triforce}. 
Specifically, TriForce, as a hierarchical framework, comprises three speculation layers, one of which relies on a standalone draft model; because no usable small draft model is available for recent RLMs, we replace that layer with a training-free NGram. 
For the middle layer, we use sliding window attention as the draft model, the same as MagicDec. 
We set $k$ equal to $1$ and $6$ for the first and middle layer, which performs best in our setup.

As shown in~\fig{fig-e2e-tps}, \sys{} consistently outperforms all baselines, achieving up to \uptocomparedtovllm{} speedup compared to \sota{} framework, vLLM. 
When compared to existing speculative decoding frameworks, \sys{} achieves up to \uptocomparedtospec{} throughput improvement. 
Interestingly, TriForce consistently falls behind MagicDec, mainly because the additional layer of NGram has a low acceptance rate (\fig{fig:fig-accu-sens} (left)) that leads to excessive computation. 
Besides, all speculative decoding methods achieve lower speedup with larger models and TP degrees. 
This is because, though available GPU memory increases, the per-token memory usage increases much less. 
For example, Qwen3-8B and -14B share the same head dimension and number of key/value heads. 
Therefore, the maximal $B$ of requests increases a lot, which approaches the saturation point $\hat{B}$, incurring more computation overhead in speculation. 

\begin{table}[!t]
\centering
\small
\caption{Breakdown of execution time (ms) on Qwen3-8B.}
\label{tab:eval-breakdown}
\begin{tabular}{lccccc}
\toprule
 & CPU & Attention & GEMM & Others & Total \\
\midrule
\textbf{vLLM} & 3.2 & 17.1 & 7.2 & 1.2 & 28.7 \\
\textbf{Ours} & 0.5 & \makecell{5.2\\ (70\%$\downarrow$)} & \makecell{8.9\\ (24\%$\uparrow$)} & \makecell{1.4\\ (17\%$\uparrow$)} & \makecell{16\\ (44\%$\downarrow$)} \\
\bottomrule
\end{tabular}
\end{table}

\MyPara{Ours vs. draft model based speculative decoding.}
We also compare \sys{} with a \sota{} draft model based speculative decoding method, EAGLE3, which require additional training.
Under this setting, \sys{} still achieves better performance over all datasets and models, despite requiring no additional training. 
\sys{} further delivers similar or higher throughput while completely avoiding the additional cost and engineering effort associated with draft-model fine-tuning and deployment.

\MyPara{Execution time breakdown.}
To breakdown the improvement, we profile the execution latency for CPU operations, attention, and GEMM in \sys{} on Qwen3-8B model with AIME dataset, and compare it with vLLM in~\reftable{tab:eval-breakdown}. 
Results show that \sys{} successfully reduces the execution time of Attention by $3.29\times$, with only a slight time increase in GEMM ($1.7$ms), consistent with our estimation in~\refsec{sec:motivation:sspec}.
Moreover, \sys{} maintains exceptionally low CPU overhead ($< 1$ ms), resulting in high GPU utilization.

\subsection{Speculation Acceptance Rate}
\label{sec:eval:acceptance_rate}
We evaluate the acceptance rate of \alg{} by measuring the average acceptance length when drafting $8$ tokens (i.e., $k=8$) across different models and datasets. 
We do not count the \textit{bonus token} for all methods.
As shown in \fig{fig:fig-accu-sens} (left), \alg{} achieves an average acceptance token length of $6.16$ out of $8$ tokens, surpassing all other drafting methods.
In comparison, both NGram and EAGLE3 can only draft fewer than $2$ accepted tokens. 
We hypothesize this is because those reasoning tasks are out-of-distribution from EAGLE3's training datasets, indicating less generality.

\subsection{Sensitivity Tests}
\begin{figure}[!t]
  \centering
  \includegraphics[width=\linewidth]{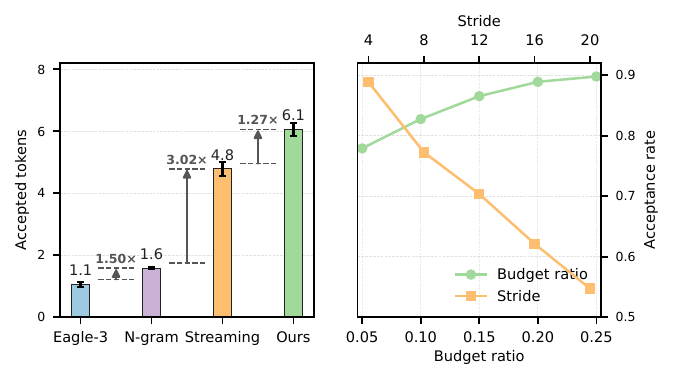}
  \vspace{-4mm}
  \caption{\textbf{Left:} Number of accepted tokens for EAGLE-3, N-gram, Streaming, and \sys{} when drafting $k{=}8$ tokens. Bars show the average across all models and datasets; error bars denote the standard deviation across model\,$\times$\,dataset combinations. \textbf{Right:} Acceptance-rate sensitivity, with budget ratio on the bottom axis ($k{=}8$) and stride on the top axis (sparsity $0.05$). Both curves are averaged over all evaluated models and datasets.}
  \vspace{-2mm}
  \label{fig:fig-accu-sens}
\end{figure}

We conduct sensitivity tests over three datasets to quantify how the speculation steps $k$ and the sparsity ratio $s$ affect the performance of \alg{}, as shown in~\fig{fig:fig-accu-sens} (right).
Despite fixing $s=0.05$ and $k=8$ in our evaluation, \sys{} has no rigid constraints on these hyperparameters during runtime, which in fact allows arbitrary combinations and heterogeneous request configurations.

\MyPara{Speculation steps $k$.}
Fixing the sparsity ratio $s=0.05$, we vary the \dl{} from $4$ to $20$.
We set the draft length $k=8$ to balance acceptance and verification overhead.

\MyPara{Sparsity ratio $s$.}
We vary the sparsity ratio in \alg{} and measure the acceptance rate when generating $8$ tokens.
We set the sparsity ratio to $0.05$, as performance saturates with further increases in selected tokens.

\subsection{Ablation Study}
\label{sec:eval:ablation}

\begin{figure}[!t]
    \centering
    \includegraphics[width=0.85\linewidth]{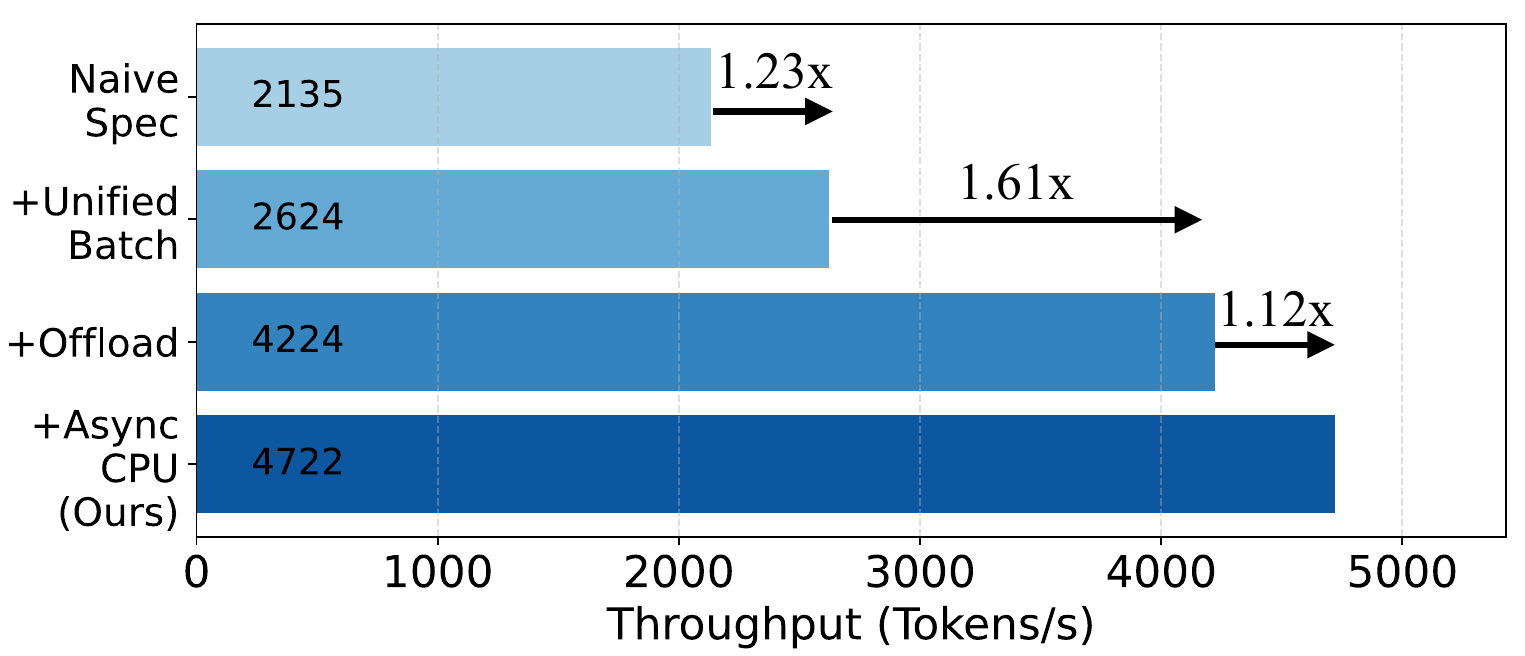}
    \caption{Ablation study of \sys{} on Qwen3-1.7B with AIME. Each component contributes to significant improvement.}
    \label{fig-eval-ablation}
\end{figure}

\begin{figure}[!t]
    \centering
    \includegraphics[width=\linewidth]{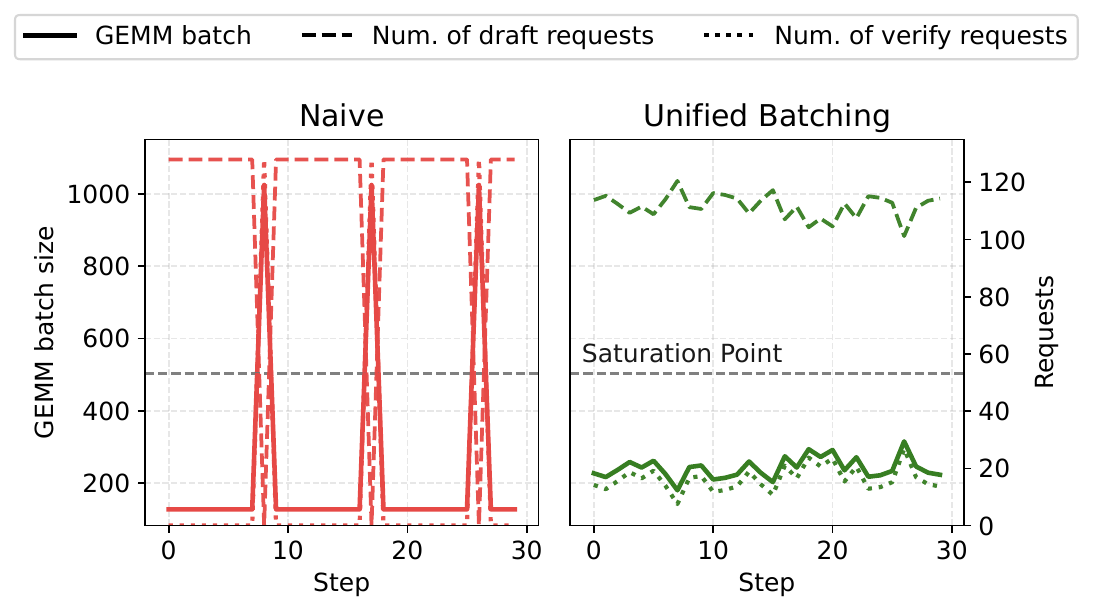}
    \caption{GEMM input batch size and request counts when running Qwen3-8B on H100 with different scheduling policies. Naive alternates between all-draft and all-verify phases, yielding fluctuation; Unified mixes draft/verify evenly with a stable batch size, preventing both GPU underutilization and oversaturation.}
    \label{fig-eval-batch}
\end{figure}

To isolate the contribution of each design component, we start from a naive implementation for \sspec{}, and incrementally enable the unified batch scheduler, \dkvm{}, and delayed verification. 
The throughput for AIME with Qwen3-8B is shown in~\fig{fig-eval-ablation}. 
Our experiments reveal that three designs boost the performance by $1.23\times$, $1.61\times$, and $1.12\times$, respectively, culminating in an aggregate throughput gain of $2.22\times$.
We also provide detailed profiling for each component: 

\MyPara{Batch size and compute utilization.}
\fig{fig-eval-batch} presents a comparative analysis of GEMM input batch sizes and hardware utilization (in TFlops) between systems with and without unified batching.
With unified batching enabled, the input batch size remains stable across iterations, while traditional scheduling causes significant variance and leads to hardware underutilization during the drafting phase.

\begin{figure}[!t]
    \centering
    \includegraphics[width=\linewidth]{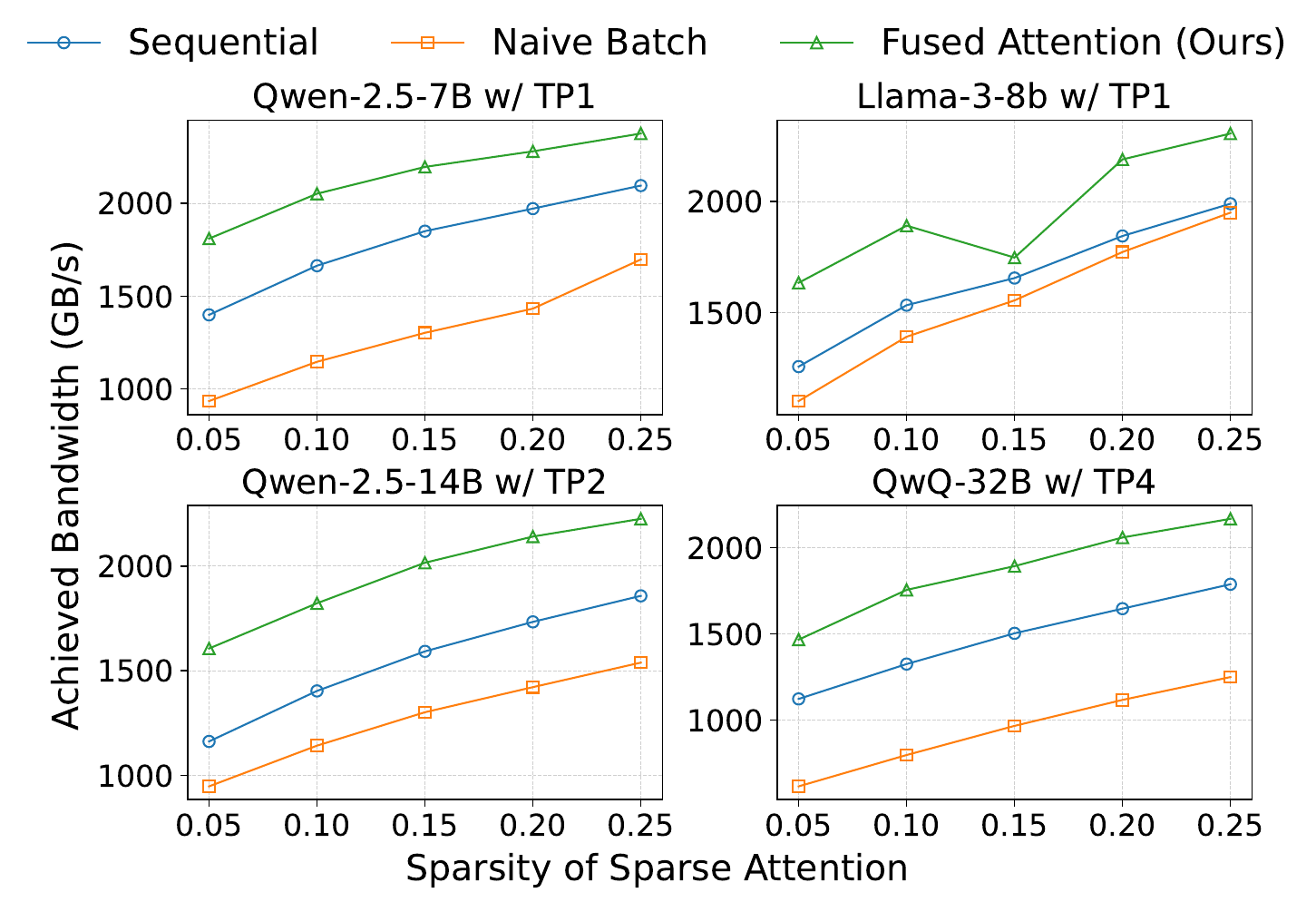}
    \caption{Performance comparison between sequentially launching FlashInfer kernels for full and sparse attention (Sequential), naively batching full and sparse attention using one FlashInfer (Naive Batch), and our fused sparse and full attention kernels.}
    \label{fig-eval-attn}
\end{figure}

\MyPara{Fused attention.}
We evaluate the performance of the fused sparse and full attention kernel against two baselines: sequentially launching two kernels (Sequential) and a naive kernel that computes both attention types jointly (Naive Batch).
The result is shown in~\fig{fig-eval-attn}.
Our fused kernel achieves a $1.3\times$ speedup over sequentially running and a $1.8\times$ speedup compared to naive batching, which comes from the following two factors:
(1) best kernel configuration dispatched to both draft and verification attention; 
(2) higher bandwidth utilization when having more transaction bytes within a single kernel to overlap pipeline latency.

\MyPara{Memory capacity utilization.}
We compare the memory utilization and recomputation of \sys{} with baselines including oracle and preemption as described in~\refsec{sec:design:offloading}.
As shown in~\fig{fig-design-offloading}, \sys{} utilizes nearly all available GPU memory without incurring recomputation.
We further quantify the overhead of offloading by comparing the execution time with offload operations enabled against a baseline where these operations are replaced with an empty kernel.
The results indicate that offloading prolongs cycle time by only $0.5$\% on average, which is practically negligible.

\section{Discussion}
\MyPara{\sys{} on mixture-of-expert (MoE) models.} 
\sys{} can be seamlessly applied to MoE models, as only the attention module is involved without modifying FFN. 
Furthermore, as only a subset of the experts is activated during inference, the input token size per expert decreases significantly, which increases the saturation point $\hat{B}$ for the sparsified FFN computation. 
Therefore, self-speculation has a higher potential due to the reduced computation overhead.

\MyPara{\sys{} with multi-token prediction (MTP).} 
\sys{} can be combined with other lightweight drafting methods, including EAGLE3 and MTP, into a hierarchical speculation approach as proposed in TriForce~\cite{sun2024triforce}. 
For example, MTP is used to draft $k_1$ tokens at once, which are verified by \alg{} into a buffer; when $k_2$ tokens are accepted and accumulated, the underlying full attention is used for verification. 
Such a hierarchical approach can reduce the amount of FFN computation besides \kvc{} loading, leading to a great opportunity for further speedup.

\MyPara{Limitation.}
The proposed method focuses on improving the memory efficiency of long-generation workloads. 
For tasks with short contexts, the maximal concurrent batch size is large enough to saturate GPU computation, making the overall workload compute-bound~\cite{zhu2024nanoflowoptimallargelanguage}. 

\section{Related Work}
\label{sec:related}

\MyPara{Long-context sparse attention.} 
Prior works have widely explored speeding up attention by exploiting its inherent sparsity. 
Static approaches~\cite{xiao2024efficientstreaminglanguagemodels,zhang2023h2oheavyhitteroracleefficient,li2024snapkvllmknowslooking} employ a fixed KV-cache pruning strategy, thus cannot adapt to evolving reasoning contexts. 
While query-aware methods~\cite{tang2024questqueryawaresparsityefficient,zhu2025tacticadaptivesparseattention,xiao2024infllmtrainingfreelongcontextextrapolation,wu2025tokenselectefficientlongcontextinference,lin2025twilight} dynamically identify important tokens during generation, they incur non-negligible overhead in estimating token importance. 
In contrast, \alg{} naturally reuses attention scores from the verification phase of speculative decoding, to dynamically select critical tokens for the next drafting phase, with minimal overhead.

\MyPara{Speculative decoding.}
Speculative decoding provides lossless acceleration for long-output tasks. Training-based approaches—including the EAGLE series~\cite{li2025eaglespeculativesamplingrequires,li2024eagle2fasterinferencelanguage,li2025eagle3scalinginferenceacceleration}, Hydra~\cite{ankner2024hydrasequentiallydependentdraftheads}, Multi-token Prediction~\cite{deepseekai2025deepseekv3technicalreport} and EESD~\cite{liu2024speculativedecodingearlyexitingfaster}—improve acceptance rates effectively via different draft model designs or learning approaches, but at the cost of higher deployment complexity. 
Training-free methods like N-gram~\cite{leviathan2023fastinferencetransformersspeculative}, Lookahead Decoding~\cite{fu2024breaksequentialdependencyllm}, and SAM Decoding~\cite{hu2024samdecodingspeculativedecoding} predict future tokens from the current context, suffering from degraded accuracy on reasoning tasks.
MagicDec~\cite{chen2024magicdec} and TriForce~\cite{sun2024triforce} adopt static sparse attention as draft for long-input scenarios but struggle with long, dynamic reasoning outputs. 
The suffix tree approach in RhymeRL~\cite{he2025history} and SuffixDecoding~\cite{oliaro2024suffixdecoding} is effective in RL rollouts but does not adapt to serving, where each document only occurs once.

In contrast, \sys{} identifies attention as the bottleneck in reasoning inference and seamlessly integrates the original model with an accurate dynamic sparse attention module \alg{} as the draft model, achieving substantial end-to-end speedups without requiring extra training or storage.

\section{Conclusion}
\label{sec:conclusion}
Due to long output sequences, reasoning model inference is heavily memory-bound.
We propose \sys{}, a lossless and training-free serving framework for reasoning models that adopts sparse self-speculation.
\sys{} identifies critical tokens during full attention in the verification phase and uses them to guide sparse attention during drafting.
With system-level optimizations—including a \ubs{}, delayed verification, and \dkvm{}—\sys{} achieves a \uptocomparedtospec{} throughput improvement over existing serving frameworks and speculative decoding baselines.
\section*{Acknowledgements}
We thank MIT-IBM Watson AI Lab, Amazon and National Science Foundation for supporting this research. We
thank NVIDIA for donating the DGX server.

\bibliographystyle{mlsys2025}
\bibliography{_main}

\appendix

\end{document}